\documentclass[11pt]{article}

\usepackage[authoryear,longnamesfirst]{natbib}
\usepackage{appendix}
\usepackage{amsmath,amssymb,amsfonts}
\usepackage[ruled]{algorithm2e}
\usepackage{graphicx}
\usepackage{subfig}
\usepackage{textcomp}
\usepackage{xcolor}
\usepackage{url}
\usepackage{array}
\usepackage{setspace}

\usepackage[margin=1in]{geometry}
    
\newcommand{\repartition}[2]{\mathbf{T}_{\{#1\}\rightarrow\{#2\}}}

\newcolumntype{"}{@{\hskip\tabcolsep\vrule width 1pt\hskip\tabcolsep}}

\title{Model-Parallel Fourier Neural Operators as Learned Surrogates for Large-Scale Parametric PDEs}

\author{
    {\small Thomas J. Grady II$^1$}\\
    \texttt{\small tgrady@gatech.edu}
    \and
    {\small Rishi Khan$^2$} \\
    \texttt{\small rishi@extreme-scale.com}
    \and
    {\small Mathias Louboutin$^1$} \\
    \texttt{\small mlouboutin3@gatech.edu}
    \and
    {\small Ziyi Yin$^1$} \\
    \texttt{\small ziyi.yin@gatech.edu}
    \and
    {\small Philipp A. Witte$^3$} \\
    \texttt{\small pwitte@microsoft.com}
    \and
    {\small Ranveer Chandra$^3$} \\
    \texttt{\small ranveer@microsoft.com}
    \and
    {\small Russell J. Hewett$^{3,\ast}$} \\
    \texttt{\small rhewett@microsoft.com}
    \and
    {\small Felix J. Herrmann$^1$} \\
    \texttt{\small felix.herrmann@gatech.edu}
}

\newcommand\blfootnote[1]{%
  \begingroup
  \renewcommand\thefootnote{}\footnote{#1}%
  \addtocounter{footnote}{-1}%
  \endgroup
}

\date{}

\begin{document}

\maketitle


%\doublespacing

\begin{abstract}
    Fourier neural operators (FNOs) are a recently introduced neural network architecture for learning solution operators of partial differential equations (PDEs), which have been shown to perform significantly better than comparable deep learning approaches. Once trained, FNOs can achieve speed-ups of multiple orders of magnitude over conventional numerical PDE solvers. However, due to the high dimensionality of their input data and network weights, FNOs have so far only been applied to two-dimensional or small three-dimensional problems. To remove this limited problem-size barrier, we propose a model-parallel version of FNOs based on domain-decomposition of both the input data and network weights. We demonstrate that our model-parallel FNO is able to predict time-varying PDE solutions of over 2.6 billion variables on Perlmutter using up to 512 A100 GPUs and show an example of training a distributed FNO on the Azure cloud for simulating multiphase CO$_2$ dynamics in the Earth's subsurface.
\end{abstract}

\section{Introduction}
\subsection{Motivation}
\label{sec:motivation}
\blfootnote{$^1$Georgia Institute of Technology, $^2$Extreme Scale Solutions, $^3$Microsoft, $^\ast$Research performed at Virginia Tech. Current affiliation Microsoft.}
Numerical simulators play an important role in many scientific fields and industries such as weather forecasting, aerodynamical design, medical, and seismic imaging or reservoir simulations \citep{gokhberg2016full, schulthess2018reflecting, louboutin2019devito, su2021min3p}. Traditional approaches to numerical simulators based on finite differences, volumes, or elements are designed to be highly accurate, meaning that errors of numerical approximations are quantifiable and numerical solutions are consistent with the original (continuous) problem formulation \citep{leveque2007finite, hughes2012finite}. In addition, traditional numerical methods are also generic, meaning that a discretized PDE can be solved for any set of initial/boundary conditions and input parameters, as long as the stability criteria of the respective discretization are met. However, these characteristics of numerical simulators come at a price, as they involve strict sets of conditions of how problems are discretized in space and time, which often leads to large, stiff systems of linear and non-linear equations that need to be solved repeatedly via expensive iterative inversion procedures \citep{burden2015numerical}. The runtime of the forward model of many simulations for real-world applications such as weather forecasting or reservoir simulations can easily lie in the range of a few hours to multiple days, which limits their applicability for problems that require a large number of simulations, such as uncertainty quantification, inverse problems, or numerical optimization.

AI-driven approaches to numerical simulations promise the possibility to train fast surrogate models for approximating solutions of partial differential equations (PDEs), which can be evaluated on the order of seconds rather than hours \citep{sirignano2018dgm, lu2019deeponet, karniadakis2021physics}. In contrast to conventional numerical solvers, which have a fixed evaluation cost, AI-driven approaches effectively front-load the computational burden to the training time (offline, including the simulation of training data), whereas at inference (online) time, trained models can be evaluated several orders of magnitude faster than the corresponding simulator. This approach is therefore beneficial for applications that require a large number of simulations, where the cost of data generation and training itself will be offset at inference time in practical situations where large numbers of simulations are required. One example for such a scenario is well location optimization in reservoir simulations, in which operators want to identify the optimal number and locations of wells for oil and gas production \citep{nasrabadi2012well}. This combinatorial optimization problem is conventionally approached with genetic optimization algorithms or more recently reinforcement learning (RL), but both approaches still require on the order of thousands of simulations and are therefore not feasible for large-scale problems \citep{onwunalu2010application, bukhamsin2010optimization}. Deep learning-based surrogate models on the other hand, can be trained at a fixed upfront cost and evaluated at a fraction of the runtime of a numerical simulator, thus making it possible to use the surrogate model during optimization \citep{wang2022efficient, salehian2022multi}. In the context of well location optimization, recent empirical results suggest the break even point (i.e. where the cost of directly running the simulator for each evaluation of the forward operator exceeds that of generating data and training a surrogate model) is on the order of a few thousand forward operator evaluations. This value is well within the number of simulations typically required when solving such a complex nonlinear optimization program, or performing statistical techniques such as Markov chain Monte Carlo sampling.

\begin{figure*}[t]
    \centering
    \includegraphics[width=0.95\textwidth]{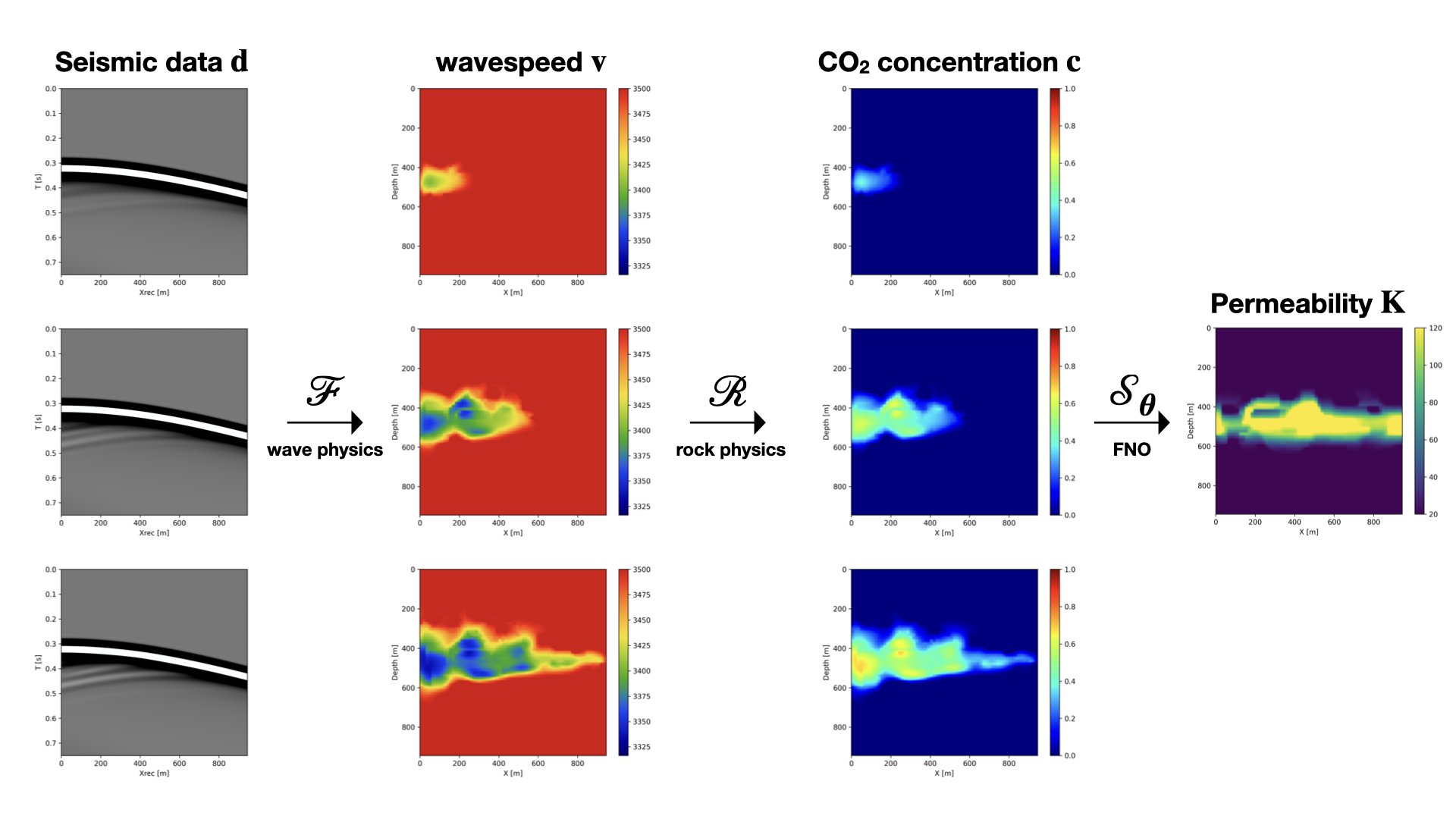}
    \caption{Coupled multi-physics inversion to estimate the subsurface permeability of a porous medium from seismic data measurements. To invert for the permeability, the authors in \citep{yin2022learned} first train an FNO that maps a permeability field to the CO$_2$ concentration history, which in turn is converted to the acoustic wave speed and used for simulating the seismic response. In the inverse problem, changes in the seismic data are first mapped to changes in the wave speed and the corresponding perturbations of the CO$_2$ concentration. Using the trained FNO, perturbations in the permeability can then be directly computed from changes in the CO$_2$ concentration via algorithmic differentiation and used to solve an inverse problem for estimating the permeability from seismic measurements. Adapted from \citep{yin2022learned}.}
    \label{fig:coupledinversion}
\end{figure*}

In addition to very fast simulation times during inference, deep learning-based approaches offer the possibility to compute gradients/sensitivities of PDEs using automatic differentiation (AD), thus making it possible to solve inverse problems without requiring users to manually differentiate the (forward) solver and implement the corresponding gradients of the simulator with respect to the input. Here, we highlight this opportunity with a recent example on subsurface CO$_2$ flow and seismic imaging (figure from \citep{yin2022learned}). The goal of this example is to estimate subsurface medium parameters such as permeability from seismic data, which is an example of a coupled multi-physics problem. Contrary to solving fluid-flow PDEs in the forward problem, the authors employ a learned Fourier neural operator (FNO; \cite{li2020fourier}) trained to predict the CO$_2$ concentration history in the subsurface from a given permeability field, which is then converted to a model of the acoustic wavespeed. The changes in wave speed induced by the expanding CO$_2$ plume can then be indirectly observed through seismic data. In the corresponding inverse problem, we are given seismic data at different points in time (i.e. during the expansion of the CO$_2$ plume), and are attempting to estimate the unknown permeability from this data (figure \ref{fig:coupledinversion}). This task is not (easily) possible if conventional simulators such as Open Porous Media (OPM) \citep{rasmussen2021open} or GEOSX \citep{gross2021geosx} are used for the CO$_2$ flow simulation, as neither framework offers sensitivities of the simulated CO$_2$ concentration with respect to the permeability. However, for an FNO implemented in deep learning frameworks like PyTorch \citep{paszke2019pytorch} or Tensorflow \citep{abadi2016tensorflow}, these sensitivities are readily available through AD, thus making it possible to implement a coupled inversion framework that enables us to directly invert for permeability from seismic data (figure  \ref{fig:coupledinversion}). 

\subsection{Challenges}
\label{sec:challenges}

One of the main challenges of adopting AI-driven solvers for real-world simulation use cases is to scale deep surrogate models such as FNOs to relevant problem sizes beyond small-scale 2D or 3D time-varying  scenarios (i.e. two or three spatial dimensions plus the time dimension). Current applications of CNNs and FNOs in the literature are based on data parallelism, where each worker gets a subset of the data and a copy of the entire model, and as such are limited to problem sizes that are supported by the amount of available memory on a single GPU \citep{yan2021robust, pathak2022fourcastnet}. I.e., with data parallelism we have to be able to store at least one full data sample (of batch size one) on a single GPU, as well as the full network and the corresponding weights, activations (hidden states) and weight gradients. For moderate 3D problem sizes beyond $64^3$ grid points, even modern GPU architectures such as the NVIDIA Ampere GPU do not provide sufficient memory to process a single training sample. To train networks for large-scale 2D and 3D time-varying problems, we are therefore required to partition the network across multiple GPUs with distributed memory. In addition to data parallelism, popular deep learning frameworks like PyTorch.Distributed \citep{paszke2019pytorch}, TensorFlow \citep{abadi2016tensorflow} or Jax \citep{frostig2018compiling} also support a model partitioning technique called pipeline parallelism, in which the layers of a neural networks are distributed across multiple GPUs. However, pipeline parallelism does not allow for arbitrary scaling, as each worker must still be able to hold the entire data and weight tensors for a single layer in memory. Other ongoing research on model parallelism has mainly focused on models for natural language processing (NLP) and especially transformer architectures, which currently represent the largest models in terms of number of parameters, such as Megatron \citep{shoeybi2019megatron} GPT3 \citep{brown2020language} or the Megatron-Turing Natural Language Generation model (NLG) \citep{smith2022using}. The latter is implemented with DeepSpeed \citep{rasley2020deepspeed}, a distributed programming framework that supports a combination of data parallelism with model/parameter parallelism (called zero-redundancy optimizer), as well as a technique that combines all of data- pipeline- and model parallelism (3D parallelism). While DeepSpeed's 3D parallelism enables users to distribute models across multiple GPUs, it does not provide users with fine-grained control over how individual tensors are partitioned (i.e. both data or weight tensors), thus making it difficult to adopt DeepSpeed for architectures such as FNOs that are distinctly different from transformers for NLP.

\subsection{Contribution}

In this work, we propose the adoption of domain decomposition for implementing model parallelism in the context of learning PDE solvers. By model parallelism via domain decomposition, we mean that we partition \textit{all} tensors of our neural network, including the input and output tensors, weight tensors, and gradient tensors along one or more of the feature dimensions (i.e. space and time). This stands in contrast to data parallelism, where tensors are only partitioned along the batch dimension (which is conventionally the first tensor dimension). Note that neural network weights do not have a batch dimension so, in data parallelism, each worker maintains a full copy of the network at all times, which becomes problematic as soon as the network does not fit onto a single GPU. In contrast, domain decomposition in principle enables us to scale to arbitrary network and data sizes, as not only the input data, but also weights and hidden states are partitioned across workers, so no single worker ever needs to store the full network. In contrast to DeepSpeed's 3D parallelism for NLP transformers, we introduce a new distinct tensor partitioning strategy for the aforementioned Fourier neural operators (FNOs) \citep{li2020fourier}, which use spectral convolutions whose weights are elementwise operators and are thus naturally model-parallel, but also require distributed multi-dimensional Fourier transforms as part of their architecture.  We base our FNO implementation on DistDL \citep{distdl_2021}, a Python package that provides domain decomposition support for PyTorch by integrating communication primitives as linear operators into PyTorch's default AD tool \texttt{autograd} \citep{hewett2020linear}. Like Mesh Tensorflow (\cite{mtf}; an extension to the Tensorflow machine learning library that adds domain decomposition to layers), DistDL also allows users to introduce domain decomposition to base layers (e.g. convolution, matrix multiplication, etc.) of its underlying neural network framework (i.e. PyTorch). However, DistDL also enables the use of data and weight tensors of arbitrary dimensionality and fine grain control of tensor partitions, both of which are critical for model parallelism in FNOs. Furthermore, DistDL also exposes its underlying parallel primitives as PyTorch \texttt{Modules}, allowing for easier development of custom domain-decomposed network layers and architectures.

\subsection{Background --- Fourier Neural Operators}

Neural operators (NOs; \cite{li2020neural}) are a recently introduced neural network architecture that learn mappings between infinite-dimensional function spaces, in contrast to traditional neural networks wherein mappings are learned between large (but ultimately finite) dimensional vector spaces. To do this, neural operators often employ architectural techniques to ensure that their output does not significantly vary with changes to the discretization of the corresponding input. In the context of learning solution operators to families of elliptic PDEs, NOs attempt to learn the mapping
\begin{equation}
    \mathcal{G}_\theta: \mathcal{A} \rightarrow \mathcal{U}
\end{equation}
where $\theta \in \Theta$ describes the parameterization of the PDE, $\mathcal{A}$  is a function space containing the initial conditions of the PDE, and $\mathcal{U}$ is a function space containing the solution of the PDE \citep{li2020fourier} \citep{li2020neural}. To learn this mapping, NOs employ an iterative architecture, constructing a sequence of functions $\nu_1, \dots, \nu_K$ in a lifted space. "Lifted" here means that the input function $a: \mathbb{R}^{d_a} \rightarrow \mathbb{R}$ is transformed to the first function in the sequence $\nu_1: \mathbb{R}^d \rightarrow \mathbb{R}, d > d_a$ via some pointwise transformation at each point $\mathbf{x} \in \mathbb{R}^{d_a}$ (e.g. in the discrete case, an affine transformation and pointwise nonlinearity along the channel dimension of the input tensor). The last value in this sequence $\nu_K$ is then projected down to an output $u: \mathbb{R}^{d_u} \rightarrow \mathbb{R}$ via a similar pointwise transformation, with the training objective that $u$ matches the solution of the PDE in a given norm (e.g. $L^2$, Sobolev). The iterative update between elements of this sequence is given by
\begin{equation}
    \nu_{k+1}(\mathbf{x}) = \sigma \left( \mathbf{W} \nu_k(\mathbf{x}) + \left(\mathcal{K}(\phi) \nu_k\right)(\mathbf{x}) \right)
\end{equation}
where $\sigma$ is a nonlinear pointwise function, $\mathbf{W}$ a learned linear transformation along the channel dimension, and $\mathcal{K}(\phi)$ a kernel integral operator with learned parameterization $\phi$. Fourier neural operators (FNOs) choose this kernel operator to be
\begin{equation}
    \left( \mathcal{K}(\phi)\nu_k \right)(\mathbf{x}) = \mathcal{F}^{-1} \left(\mathbf{R}_\phi \cdot \left( \mathcal{F}\nu_k \right) \right)(\mathbf{x}),
\end{equation}
where $\mathcal{F}$ is the Fourier transform, and $\mathbf{R}_\phi$ is a restriction operator, which contains a low-pass filter and learned pointwise weight multiplication parameterized by $\phi$ \citep{li2020fourier}. This operator is referred to as a \textit{spectral convolution}. The cutoff of the low pass filter in $\mathbf{R}_\phi$ will depend on a user-defined parameter describing how many Fourier-modes to keep in each dimension.

When learning solutions to time-dependent PDEs, FNOs are most often trained on discretized pairs of input data $(\mathbf{X},\mathbf{Y})$, where $\mathbf{X}$ is a multidimensional tensor containing a discretization of the initial state $a(\mathbf{x}) \in \mathcal{A}$ and $\mathbf{Y}$ is a discretization of the time-evolving solution to the PDE, $u(\mathbf{x}, t) \in \mathcal{U}$.

\newpage

\section{Parallel Implementation}

\subsection{Background - Abstraction of Parallelism} \label{sec:parallel_abstraction}
In order to successfully implement complex parallel algorithms acting on high-dimensional tensors within the context of an automatic differentiation framework (e.g. PyTorch \texttt{autograd} \citep{paszke2017automatic}), it is important to have a clear and expressive abstraction for high-dimensional parallel programming primitives and their adjoints \citep{utke2009toward}. Specifically, in the context of neural networks, domain decomposition of data and network weights poses a complex engineering challenge. Unlike in traditional PDE solvers where the domain of interest is of fixed size and decomposed over a fixed worker topology from one timestep to the next (e.g. Devito \citep{louboutin2019devito}), neural networks often deal with a sequence of transformations where the shape, dimensionality, and decomposition of the data and weight tensors may vary from layer to layer. Thusly, they have challenges more akin to those of domain-decomposed PDE solvers with adaptive mesh refinement (e.g. PARAMESH \citep{macneice2000paramesh}). To deal with this complexity, we follow the approach of \cite{hewett2020linear}, describing parallel primitives in terms of linear operators acting on domain-decomposed tensors. In \cite{hewett2020linear}, the authors derive a rigorous definition of these primitives on distributed memory supercomputers from only very basic memory operations such as clearing, copying, and moving data. Here, we avoid a full re-derivation and focus only on the background information and operators necessary to implement a distributed FNO. We denote the distribution of a tensor across a Cartesian topology of parallel workers as a \textit{partition}. Note that this abstraction makes no assumptions about the underlying device, data type, or other properties of the algorithm, and describes only the pattern of communication between parallel workers.

\begin{figure}[h]
    \centering
    \includegraphics[width=0.6\textwidth]{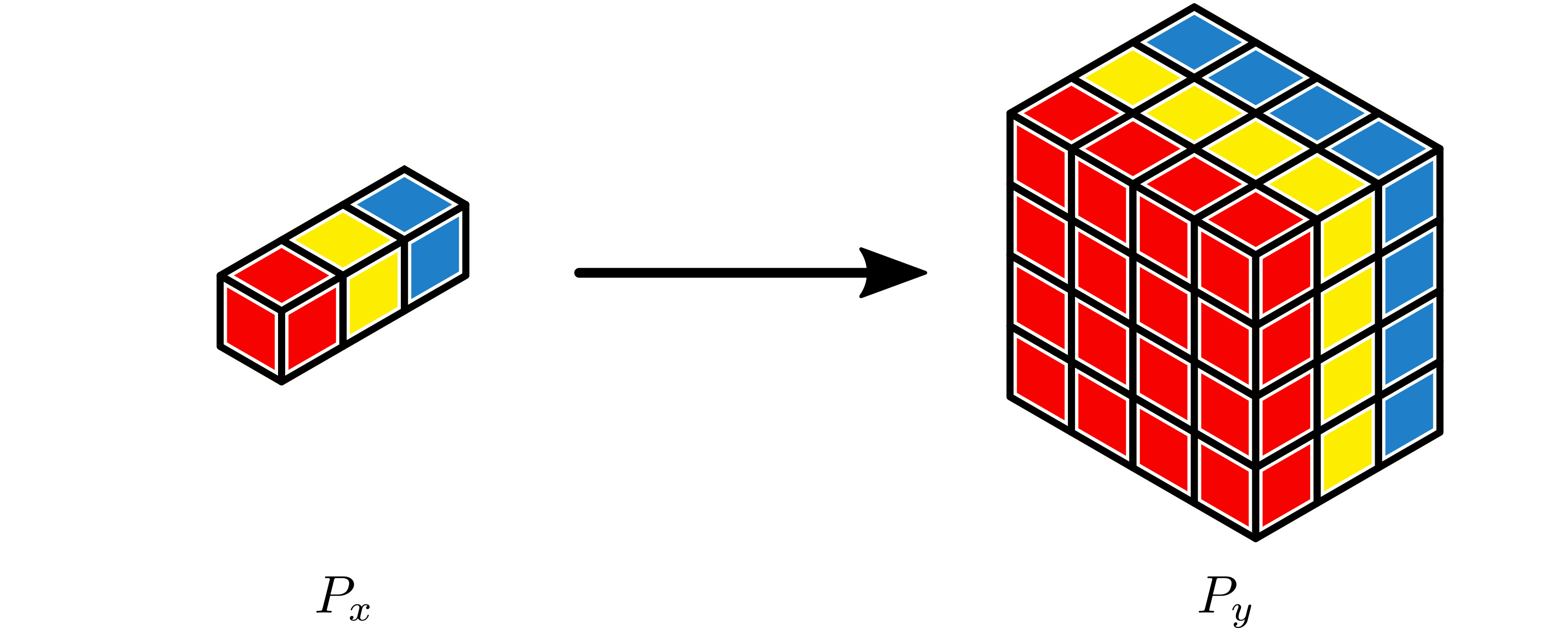}
    \caption{Broadcast of a tensor distributed over a $1 \times 1 \times 3$ partition $P_x$ to a $4 \times 4 \times 3$ partition $P_y$. Note that following DistDL broadcasting rules, the global tensor maintains the same size in the third dimension, as $P_x$ and $P_y$ are of equal shape in that dimension. Figure from \citep{distdl_2021}.}
    \label{fig:bcast}
\end{figure}

To implement a distributed FNO, two core parallel primitives are required. The first of these primitives is \textit{broadcast}, which copies subtensors of a tensor partitioned on one set of workers to another. As described in \cite{hewett2020linear}, the broadcast operation on tensors extends beyond the classical parallel broadcast primitive.  While it can trivially represent the classical operation, i.e., copying data from one worker to many, it can also represent an extension of this operation to tensor partitions. As long as the input and output partitions satisfy the DistDL broadcasting rules (a subset of the NumPy broadcasing rules \citep{harris2020array}), the action of this operator between two partitions $P_x$ and $P_y$, $\mathbf{B}_{\{P_x\}\rightarrow\{P_y\}}$, will copy an input tensor $\mathbf{x}$ along the appropriate dimensions as seen in figure \ref{fig:bcast}.
Following the definition of the adjoint, a broadcast in the forward evaluation will induce a sum-reduction in the gradient calculation.

\begin{figure}[h]
    \centering
    \includegraphics[width=0.6\textwidth]{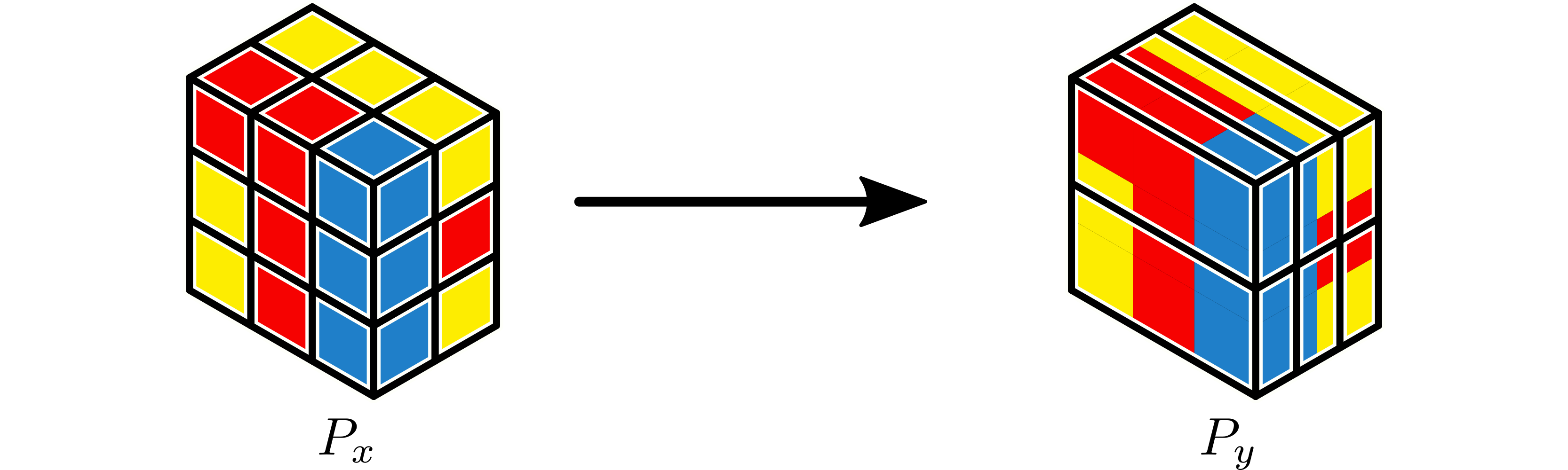}
    \caption{Repartition operator changing the partition of a tensor from a partition $P_x$ of shape $3 \times 3 \times 2$ to a partition $P_y$ of shape $1 \times 2 \times 3$. Figure from \citep{distdl_2021}.}
    \label{fig:repartition}
\end{figure}

The second of these primitives is \textit{repartition}, a high-dimensional generalization of \textit{all-to-all}. The action of this operator, $\repartition{P}{Q}$, changes the distribution of the data from one partition $P$ to another partition $Q$,  as demonstrated in figure \ref{fig:repartition}.  While $P$ and $Q$ are not required to have the same number of workers, they are required to have the same number of dimensions as the input tensor.  In higher dimensions, an implementation of this primitive is not trivial as any worker may need to send or receive subtensors to or from any or all other workers (i.e., a many-to-many operation).
Repartition is in some sense the most ``general'' possible communication primitive for tensors and its adjoint is also a repartitioning, namely from $Q$ to $P$.

Ultimately, this linear algebraic formulation of parallel primitives for manipulating domain-decomposed tensors allows for nontrivial parallel operations to be cleanly expressed within the mathematical formulation of FNOs, and the usage of DistDL allows for easy integration of these primitives within PyTorch network architectures.

\subsection{Pointwise Affine Transformations}\label{sec:paff}
\begin{figure}
    \centering
    \includegraphics[width=0.6\textwidth]{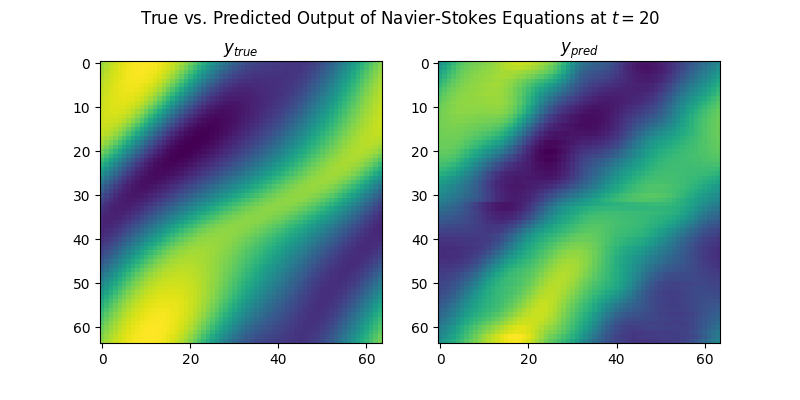}
    \caption{Artifacting at parallel worker boundaries and solution inaccuracies on the validation dataset of a fluid flow problem caused by training a distributed FNO over a $2 \times 2$ partition without broadcasting the weight matrices and bias vectors before applying pointwise transformations.}
    \label{fig:artifacts}
\end{figure}

One of the two core components of FNOs that must be adapted for a distributed setting are the affine transformations along particular dimensions
\begin{equation}
    \mathbf{y} = \mathbf{Wx} + \mathbf{b},
    \label{eqn:affine}
\end{equation}
where the action of $\mathbf{W}$ on $\mathbf{x}$ can be thought of as a tensor contraction along a given dimension of $\mathbf{x}$. In our implementation, we make the assumption that all pointwise affine transformations act only along dimensions of $\mathbf{x}$ which are not distributed (i.e. the shape of the partition of $\mathbf{x}$ in that dimension is 1). In the context of FNOs, this is a safe assumption to make, as there are no extremely large weight matrices used in these transforms which would necessitate a distributed matrix-tensor contraction. Assuming $\mathbf{W}$ and $\mathbf{b}$ are stored on a size 1 partition $P_r$, we can apply the broadcast operator from $P_r$ to a partition $P_d$ of size 1 in dimension $d$ to both $\mathbf{W}$ and $\mathbf{b}$. Following the broadcasting rules described in section \ref{sec:parallel_abstraction}, $\mathbf{W}$ and $\mathbf{b}$ will be identical on all workers, meaning each worker can compute $\mathbf{Wx} + \mathbf{b}$ along dimension $d$ locally. This gives the operator
\begin{equation}
    \mathbf{y} = (\mathbf{B}_{\{P_r\}\rightarrow\{P_d\}}\mathbf{W})\mathbf{x} + \mathbf{B}_{\{P_r\}\rightarrow\{P_d\}}\mathbf{b}.
    \label{eqn:affine_bcast}
\end{equation}
As opposed to each worker having its own independent weight and bias tensors, the inclusion of a broadcast of the weight and bias terms ensures no solution discontinuities occur at worker boundaries as seen in the center of the output in figure \ref{fig:artifacts}.
As an example, consider a tensor $x$ of shape $10 \times 20 \times 64 \times 64$ which is distributed over a partition $P_x$ of shape $1 \times 2 \times 2 \times 2$. To apply a $20 \times 20$ weight matrix $\mathbf{W}$ along the channel (2nd) dimension of $\mathbf{x}$, we first apply a repartition operator $\repartition{P_x}{P_c}$ to $\mathbf{x}$ where $P_c$ is of shape $1 \times 1 \times 4 \times 2$ (ensuring each worker's subtensor of $\mathbf{x}$ has size 20 in the channel dimension). We then broadcast $\mathbf{W}$ from a $1 \times 1 \times 1 \times 1$ partition $P_r$ to $P_c$ using the broadcast operator $\mathbf{B}_{\{P_r\} \rightarrow \{P_c\}}$, and locally compute $\mathbf{Wx}$ on each worker.

\subsection{Distributed Fourier Transform}
Performing fast discrete Fourier transforms (FFTs) on domain-decomposed tensors is a well-studied problem. Current state-of-the-art approaches \citep{dalcin2019fast, pippig2013pfft} use an iterative procedure, wherein a chain of repartition operators and sequential FFTs are used to compute the entire FFT over distributed data without ever having to compute a parallel FFT along any dimension. Note that this works only because the Fourier transform is separable. From a linear algebra viewpoint, the distributed fast Fourier transform (DFFT) $\mathcal{F}_{\sf dist}$ of an $n$-dimensional tensor $\mathbf{x}$ distributed over a partition $P$ can then be written\footnote{Optionally, there is an additional repartition operator $\repartition{P_{\mathcal{I}_k}}{Q}$ to change the partition of the data to some output partition $Q$, but due to the structure of FNOs, this step is not performed.}
\begin{equation}
    \mathcal{F}_{\sf dist}\mathbf{x} =
    \mathcal{F}_{\mathcal{I}_k}\repartition{P_{\mathcal{I}_{k-1}}}{P_{\mathcal{I}_{k}}}
    \dots
    \mathcal{F}_{\mathcal{I}_1}\repartition{P}{P_{\mathcal{I}_1}}\mathbf{x}
\end{equation}
\noindent where 
\begin{equation}
    \bigcup_{j=1}^k \mathcal{I}_j = \{1, 2, \dots, n\}, \quad \mathcal{I}_{j_1} \cap \mathcal{I}_{j_2} = \emptyset \ \forall \ j_1 \neq j_2. \\
\end{equation}
\noindent Each $\mathcal{I}_j$ describes an index set of dimensions over which to apply the sequential multidimensional Fourier transform $\mathcal{F}_{\mathcal{I}_j}$, and denotes which dimensions of $\mathbf{x}$ are present in their entirety on each worker (i.e. the shape of $P_{\mathcal{I}_j}$ is 1 in those dimensions). Combining this formulation with PyTorch's native multidimensional FFT yields a clean and powerful implementation of a distributed, differentiable $n$-dimensional FFT, which can also be further generalized to other combinations of separable transforms. Choosing this selection of index sets is generally problem specific, but a good rule of thumb is to select them such that the number of repartition operators (and thus generalized all-to-all communications) is minimized while still being able to fit all subtensors within the memory constraints of their corresponding parallel workers at all steps of the DFFT.  In our implementation, we find it necessary to only apply a single repartition operator between sequential transforms. I.e. For an $n$-dimensional input first taking an FFT over the first $\frac{n}{2}$ dimensions, repartitioning, and then taking an FFT over the last $\frac{n}{2}$ dimensions.
We can then take this DFFT and simply replace the standard FFT in the spectral convolution to get the a mathematical representation of distributed spectral convolution operator
\begin{equation}
    \left(\mathcal{S}_{\sf dist}\nu_k\right)(\mathbf{x}) = \left(\mathcal{F}_{\sf dist}^\top(\mathbf{R}_\phi \cdot (\mathcal{F}_{\sf dist}\nu_k)) \right)(\mathbf{x}).
    \label{eqn:sconv_dist}
\end{equation}

\begin{figure}[h]
    \centering
    \includegraphics[width=0.5\textwidth]{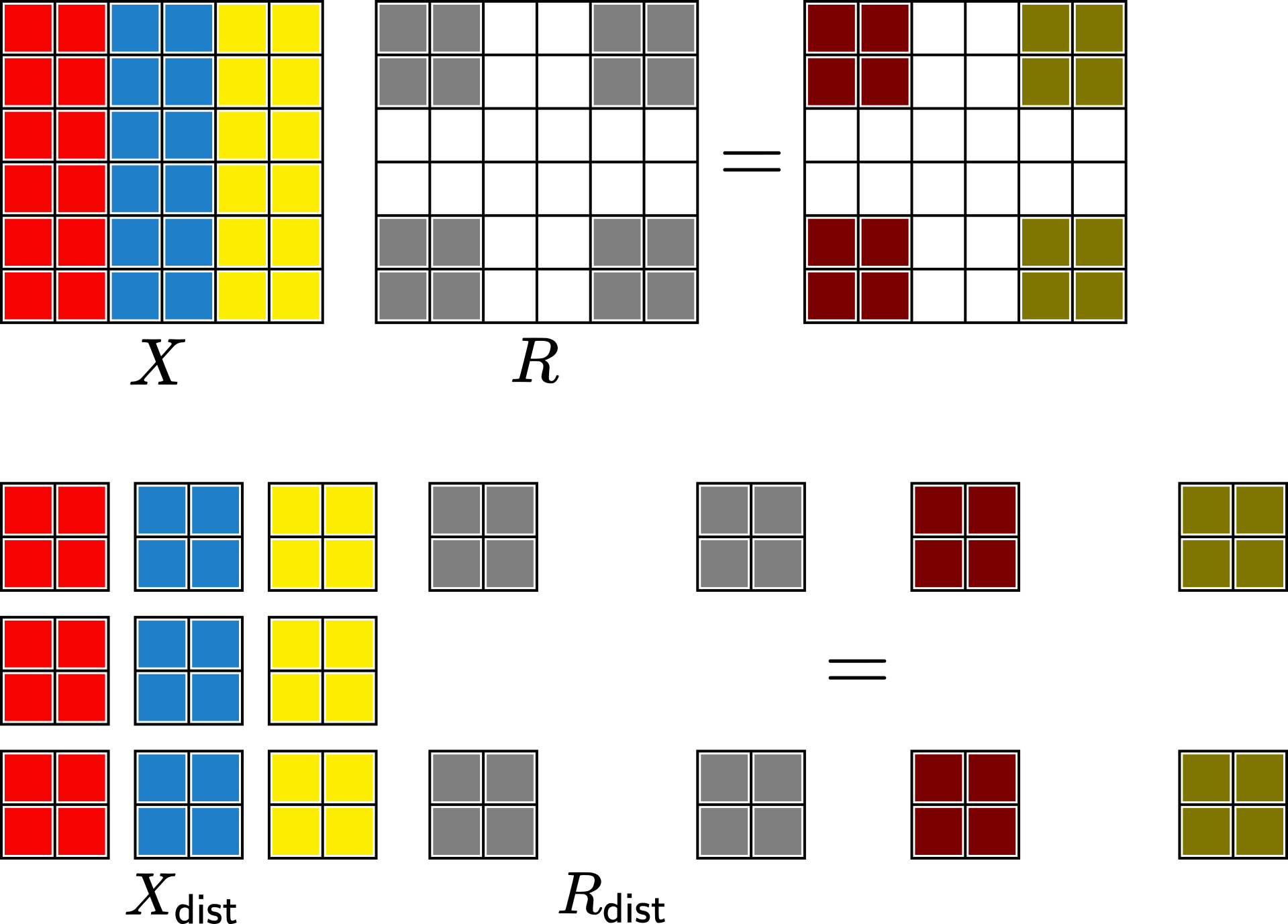}
    \caption{Sequential versus distributed application of $\mathbf{R}_\phi$ on 2D data distributed over a $3 \times 3$ partition.}
    \label{fig:R_phi}
\end{figure}

Practical considerations require that only workers containing nonzero values in the Fourier domain after the application of the low-pass filter  within $\mathbf{R}_\phi$ perform any work. To achieve this, each worker uses information about the underlying partition of the data and the location of its corresponding local subtensor in the global distributed tensor $\mathbf{x}$ to compute whether it will contain a nonzero value after application of $\mathbf{R}_\phi$. If so, it applies the restriction and pointwise weight multiplication, otherwise the value is multiplied by zero. Note that all of this is happening on the output partition $P_{\mathcal{I}_k}$ of the forward DFFT $\mathcal{F}_{\sf dist}$, as to remove the need for an unnecessary all-to-all communication to get the data back to the original partition before applying $\mathbf{R}_\phi$. See figure \ref{fig:R_phi} for an illustration of an application of $\mathbf{R}_\phi$ in a simple 2D case.

\subsection{Full Network}
Combining the broadcast operator, repartition operator, and DFFT, we are now able to describe the distributed FNO (DFNO) in its entirety. Recall that the objective of an FNO acting as a surrogate to a PDE solver is to map a parameterized input function $a(\mathbf{x}; \theta)$ to a time-varying solution $u(\mathbf{x}, t)$. FNOs traditionally consist of the following sequence of transformations. First, the input function is projected to the correct number of output timesteps and lifted to a higher-dimensional space via two affine transformations along the time and channel dimensions
\begin{align}
    a_1(\mathbf{x}, t) &= (\mathbf{W}_ta + \mathbf{b}_t)(\mathbf{x}, t) \\
    \nu_1(\mathbf{x}, t) &= (\mathbf{W}_ca_1 + \mathbf{b}_c)(\mathbf{x}, t)
\end{align}
respectively. For clarity here we omit the parameter $\theta$ and assume $a$ has a time dimension of size $1$ along which $\mathbf{W}_t$ can be applied. To distribute these two transformations, we apply repartition operators before each to ensure the dimensions on which they act are present entirely on each worker as in section \ref{sec:paff}. These partitions are denoted $P_t$ and $P_c$ respectively.
\begin{figure}[h!]
    \centering
    \includegraphics[width=1.0\textwidth]{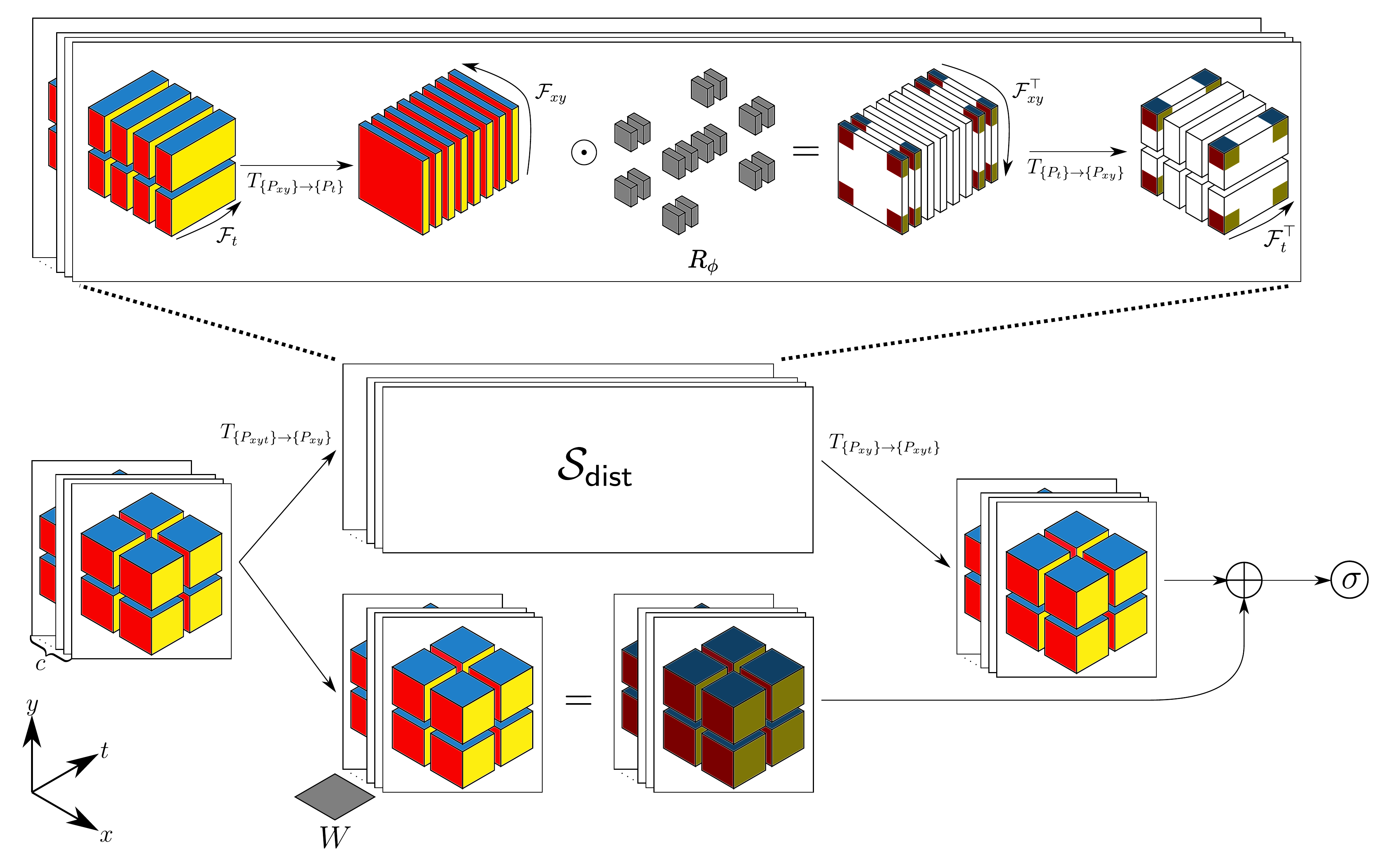}
    \caption{Distributed FNO block acting on a 2D time-varying input tensor (i.e. a 3D volume with two spatial and a time dimension), illustrating the iterative update in equation \ref{eqn:dist_block}. In this example, a tensor of shape $c \times n_x \times n_y \times n_t$ is initially distributed over a partition $P_{xyt}$ of shape $1 \times 2 \times 2 \times 2$, indicating that it is partitioned over 8 workers in the $x, y$, and $t$ dimensions. Here, $c$ is the channel dimension and is illustrated via stacked planes, and $n_x, n_y,$ and $n_t$ are the size of the input tensor in the $x$, $y$, and $t$ dimensions respectively. To apply the distributed spectral convolution operator $\mathcal{S}_{\sf dist}$ from equation \ref{eqn:sconv_dist}, a repartition operator is used to take the data to a partition of only the $x$ and $y$ dimensions, denoted $P_{xy}$. After this, a DFFT is performed by first taking an FFT along time, followed by a repartition to a partition of only the time dimension (denoted $P_t$), followed by a 2D FFT along the $x$ and $y$ dimensions. The parameterized (learned) weight/restriction tensor $\mathbf{R}_\phi$ is then multiplied with the FFT output. This value is then passed through the adjoint sequence of transformations, where $\mathcal{F}^\top$ denotes an adjoint (inverse) DFFT. As in the original implementation by \cite{li2020fourier}, $\mathbf{R}_\phi$ is sparse, containing nonzero elements only in the low-frequency modes. Below $\mathcal{S}_{\sf dist}$, the weight tensor $\mathbf{W}$  (broadcasted over $P_{xyt}$) is multiplied along the channel dimension of the input. The result of this weight multiplication and the output of $\mathcal{S}_{\sf dist}$ are then added together elementwise and passed through an elementwise nonlinear activation function $\sigma$.}
    \label{fig:dfno_block}
\end{figure}
\clearpage
We then apply the broadcasted affine operator as in equation \ref{eqn:affine_bcast}. Assuming that each weight tensor and bias vector is stored on the same size 1 root partition $P_r$, this is written as the sequence of transformations
\begin{align}
    \mathbf{W}_t^{bc} &= \mathbf{B}_{\{P_r\}\rightarrow\{P_t\}}\mathbf{W}_t \\
    \mathbf{W}_c^{bc} &= \mathbf{B}_{\{P_r\}\rightarrow\{P_c\}}\mathbf{W}_c \\
    \mathbf{b}_t^{bc} &= \mathbf{B}_{\{P_r\}\rightarrow\{P_t\}}\mathbf{b}_t \\
    \mathbf{b}_c^{bc} &= \mathbf{B}_{\{P_r\}\rightarrow\{P_c\}}\mathbf{b}_c \\
    a_1 &= (\mathbf{W}_t^{bc} \repartition{P_x}{P_t}a + \mathbf{b}_t^{bc})(\mathbf{x}, t) \\
    \nu_0 &= (\mathbf{W}_c^{bc}\repartition{P_t}{P_c} a_1 + \mathbf{b}_c^{bc})(\mathbf{x}, t),
    \label{eqn:lift}
\end{align}
where $W_t, b_t$ and $W_c, b_c$ are the weights and biases for each affine transformation respectively.
The FNO architecture then calls for a sequence of $K$ blocks consisting of a linear transformation along the channel dimension added with a spectral convolution and passed through a pointwise nonlinearity. A single iterative update in this sequence reads
\begin{equation}
    \nu_{k+1}(\mathbf{x}, t) = \sigma\left(\mathbf{W}\nu_k + \mathcal{S}\nu_k \right)(\mathbf{x}, t),
\end{equation}
where $k = 1,\dots,K$ and $\sigma$ is a pointwise nonlinear activation function. Applying the same procedure as in equation \ref{eqn:affine_bcast} for distributing pointwise linear transformations then gives the sequence of transformations for the DFNO block
\begin{align}
    \mathbf{W}^{bc} &= \mathbf{B}_{\{P_r\}\rightarrow\{P_x\}}\mathbf{W} \\
    \nu_{k+1}(\mathbf{x}, t) &= \sigma \left(\mathbf{W}^{bc}_c\nu_k + \mathcal{S}_{\sf dist}\nu_k \right)(\mathbf{x}, t),
    \label{eqn:dist_block}
\end{align}
where $x$ is the distributed tensor and $\mathcal{S}_{\sf dist}$ is the distributed spectral convolution operator derived in equation \ref{eqn:sconv_dist}. Equation \ref{eqn:dist_block} constitutes the most important component of the distributed FNO, and as such an illustration is provided in figure \ref{fig:dfno_block}.

Finally, a linear operator is used to project the output to the correct number of channels. Using a similar derivation to equation \ref{eqn:lift}, the distributed variant is
\begin{equation}
    u(\mathbf{x}, t) = \left(\mathbf{W}^{bc}_c\repartition{P_x}{P_c}\nu_K\right)(\mathbf{x}, t),
\end{equation}
where $u(\mathbf{x}, t)$ is the approximate solution of the PDE given by the operator learned by the distributed FNO, and is distributed over the partition $P_c$.

\section{Experimental Results}
\subsection{Training} \label{sec:training}

The ability to scale FNOs with domain decomposition to large problem sizes opens up the possibility to apply them to real-world simulation use cases. To showcase the value of being able to train model-parallel networks for solving large-scale PDEs, we train a distributed FNO to simulate subsurface CO$_2$ flow by solving the 3D time-varying two-phase flow equations \citep{rasmussen2021open, gross2021geosx}. Simulating CO$_2$ flow in porous media plays an important role in carbon capture and storage (CCS), where simulations are required to optimize the CO$_2$ injection location and verify that CO$_2$ does not leak from the storage site \citep{gibbins2008carbon, ringrose2020store}. A variety of deep-learning based approaches, including FNOs, have been proposed for simulating subsurface CO$_2$ flow, but so far current examples in the literature are limited to either 2D or small to medium-scale 3D time-varying problems \citep{wen2021u, yan2021robust, tang2021deep}. On the other hand, our model-parallel FNO can scale to realistic 3D problem sizes using domain decomposition across multiple GPUs, including those commonly found on cloud computing services such as the 16GB variant of the NVIDIA V100.

We train an FNO to predict the evolution of a 3D subsurface CO$_2$ plume over a given number of time steps $n_t$. The input to the network $\mathcal{K}(\mathbf{x})$ is a tensor containing both the permeability and topography (i.e. column-wise vertical displacement) at each 3D spatial position $\mathbf{x}$. In contrast to the original FNO from \cite{li2020fourier}, our input does not include the first $10$ time steps of the predicted saturation history, which is zero almost everywhere. We train our model on the CO$_2$ simulation dataset from \cite{wittesleipner2021}, which was derived from a subregion of the Sleipner benchmark \citep{sleipner2019}, a reservoir simulation model from the world's first industrial-scale CO$_2$ injection site off the coast of Norway \citep{andrew2015sleipner, furre201720}. The training dataset consists of $1000$ permeability models that were randomly generated in analogy to the original benchmark. Open Porous Media (OPM; \cite{rasmussen2021open}) was then used to simulate the corresponding saturation histories for each sample. OPM solves a large set of partial differential equations with complex boundary conditions to accurately predict the time-evolution of subsurface CO$_2$. Each individual input sample to the network has a shape of $batch \times 2 \times 60 \times 60 \times 64 \times 1$ (NCXYZT), where permeability and (geographical) topography are provided as the two input channels. The output has a shape of $batch \times 1 \times 60 \times 60 \times 64 \times n_t$. As in \cite{pathak2022fourcastnet} each training sample is centered in a volume large enough that the CO$_2$ saturation never reaches the boundary of the simulation domain.
The architecture of our network is identical to the original proposed by \cite{li2020fourier}, consisting of a total of 4 spectral convolution blocks, and a lifted space dimension of $20$. The network is trained for $30$ epochs using $800$ training data points and $100$ validation data points with a batch size of 1. When training our network, we use the Adam optimizer \citep{kingma2014adam} with a learning rate of $10^{-3}$. As in the original FNO paper, we measure the data misfit using the relative $L^p$ loss with $p=2$ \citep{li2020fourier}:
$$L(\mathbf{y}, \mathbf{\hat{y}}) = \frac{\| \mathbf{y}- \mathbf{\hat{y}} \|_p}{\|\mathbf{\hat{y}}\|_p}.$$
We train our model on a \texttt{Standard\_NC24} virtual machine on the Azure cloud, which has 4 Nvidia Volta V100 GPUs with 16 GB memory each. We therefore create a worker partition of shape $1 \times 1 \times 1 \times 4 \times 1 \times 1$, which results in an input shape per worker of $batch \times 2 \times 60 \times 15 \times 64 \times n_t$. It should be noted that this training setup does not fit within the memory of a single V100 GPU and thus constitutes true model-parallel training with domain decomposition. While our FNO implementation is in principle able to handle much larger data sizes, there are currently no larger public datasets for training multiphase flow simulators available. The training dataset in our experiment is stored in Azure's object store (Blob storage) in the Zarr format, whose corresponding Python packages provides an API for storing chunked $n$-dimensional tensors on both file systems and object stores \citep{zarr2021}. During training, each MPI rank reads its corresponding domain of the input data directly from the object store, so no single GPU has to fit the full input (or output) data into its memory at any given time.

\begin{figure}[h]
    \centering
    \includegraphics[width=0.5\textwidth]{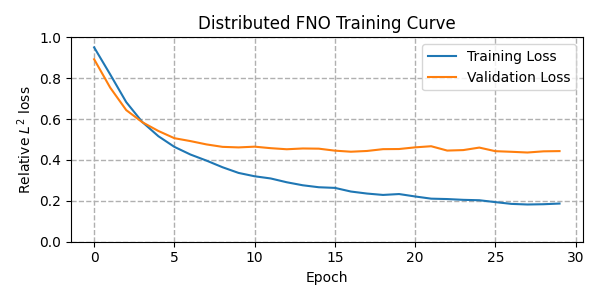}
    \caption{Training and validation loss curves for running distributed FNO training experiment. Validation loss plateaus around a value of 0.4, so the network training was halted after 30 epochs.}
    \label{fig:loss}
\end{figure}

\begin{figure*}[h]
    \centering
    \subfloat[\centering Horizontal ($x,y$) slices]{{\includegraphics[width=0.45\textwidth]{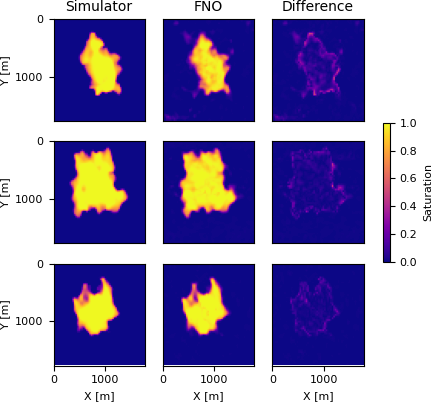} }}%
    \qquad
    \subfloat[\centering Vertical ($x,z$) slices]{{\includegraphics[width=0.45\textwidth]{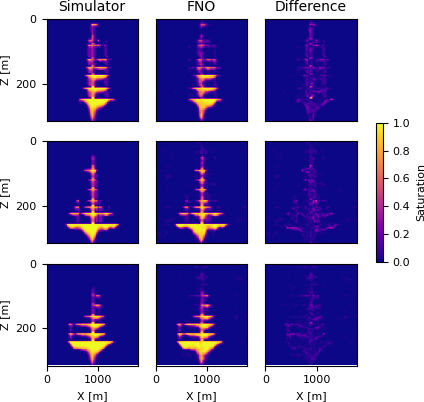} }}%
    \caption{Predicted outputs of our 4D (3D time-varying) FNO on 3 validation examples. Images shown are slices of the output 3D volume at the final timestep $t=30$, but the network is trained on and produces an output of all timesteps simultaneously. The network shows good generalization results and produces error small enough to be useful for practical applications.}%
    \label{fig:sleipner_slices}
\end{figure*}

\begin{figure}[h]
\centering
\subfloat{\label{fig:perm3d}}{\includegraphics[width=0.4\textwidth]{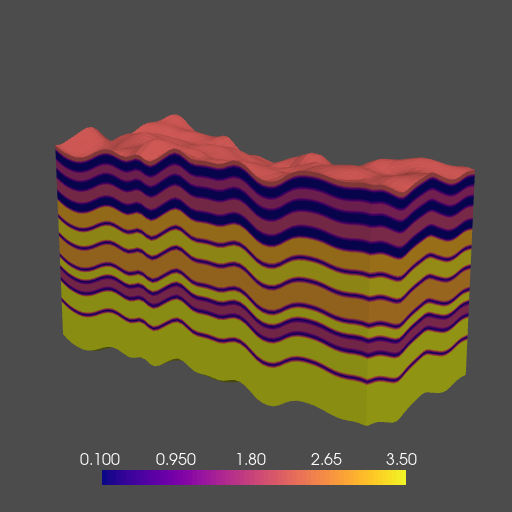}}
\hspace*{.1cm}
\subfloat{\label{fig:sat3d}}{\includegraphics[width=0.4\textwidth]{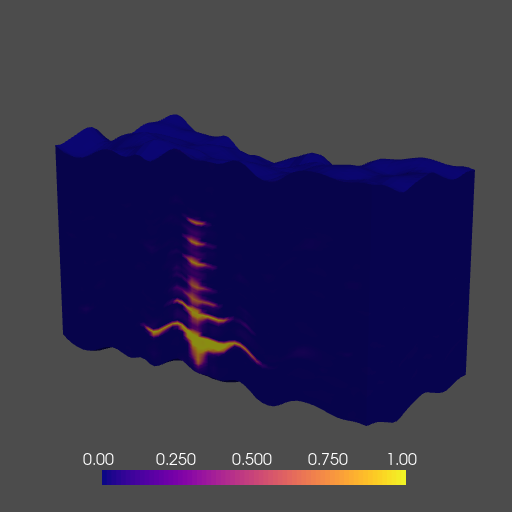}}
\caption{Input permeability/topography map and output CO$_2$ plume at the final timestep for a validation sample run through our trained 4D two-phase flow FNO.}
\label{fig:sleipner_3d}
\end{figure}

Figure \ref{fig:loss} shows the history of the training and validation loss as a function of the training epoch. It is noticeable that the validation loss stalls around a value of 0.4, while the training loss decreases until the final epoch. Nevertheless, the trained network performs reasonably well on unseen test samples, as  shown in figure \ref{fig:sleipner_slices} by a comparison of the predicted CO$_2$ saturation with the corresponding simulated data samples. The network performs well enough that its output could be used by an industry professional to make decisions about, for example, CO$_2$ storage site location, or to solve very large multi-physics inverse problems as in \citep{yin2022learned}. A three-dimensional plot of a test samples that correctly visualizes the varying grid topography is shown in figure \ref{fig:sleipner_3d}.

\clearpage

\subsection{Inference}
After paying the large upfront cost of training a distributed FNO, practitioners can then take the network and utilize its accelerated inference capabilities to solve novel and difficult problems. FNOs in particular have shown much promise in this area, reporting speedups of several orders of magnitude at inference time versus traditional numerical simulators \citep{li2020fourier}. To perform a comparison to standard two-phase flow simulators, we benchmark our distributed network versus OPM \citep{rasmussen2021open} on two different problem sizes. The first of these problems is of size $60 \times 60 \times 64$ in the spatial dimensions with 30 timesteps, the same setup as the dataset of \cite{wittesleipner2021} and the subsequent training example in section \ref{sec:training}. The second is a problem of size $68 \times 118 \times 263$ with 16 timesteps. This is the size of the Sleipner benchmark model \citep{sleipner2019}, which represents one of the smallest industry-scale models of interest to CCS practitioners. Because OPM does not currently support GPU acceleration, both it and the distributed FNO were run on a CPU node, namely an HB120rs v2 node on Azure. This node has 120 AMD EPYC 7742 CPUs with 4GB of memory per CPU. Both the simulator and FNO used 30 MPI workers with 4 OpenMP threads per worker. Table \ref{tab:speedup} shows the timing comparison.

\begin{table}[h]
    \centering
    \begin{tabular}{|c|c|c|l|}\hline
        Problem Size & OPM Time (s) & FNO Time (s) & Speedup\\\hline
        $60 \times 60 \times 64 \times 30$ & 312 & 1.15 & 271x \\\hline
        $68 \times 118 \times 263 \times 16$ & 8291 & 5.98 & 1386x \\\hline
    \end{tabular}
    \caption{Timing comparison of OPM simulator and distributed FNO at inference time. The problem size is shown as $n_x \times n_y \times n_z \times n_t$, and the speedup reported for the FNO is that versus OPM run on the same problem size with the same MPI configuration.}
    \label{tab:speedup}
\end{table}

We note that our network greatly outperforms the simulator in both cases, with the speedup growing as a function of the size of the problem and reaching a maximum value of 1386x faster than OPM on the Sleipner benchmark model. Furthermore, as the problem size grows (4.9x), the FNO time grows roughly linearly with problem size (5.2x) wheres the OPM simulation time grows much faster (26.6x).

\subsection{Scaling} \label{sec:scaling}

While training and inference using FNOs on large 3D time-varying datasets is novel in its own right, the ultimate aim of surrogate modeling for CCS or other large scientific modeling problems is to quickly be able to perform simulations that are accurate enough to be useful for previously intractable problems such as large-scale Bayesian inference, ideally using data and models too large to fit on a single computational node. Here we demonstrate the unique ability of our domain-decomposition approach to model parallelism in FNOs to achieve this through a weak scaling study and show, to our knowledge, the largest inference and gradient computations done using an FNO to date, achieving a maximum problem size\footnote{Due to limitations on system resource usage we were unable to scale past this size, but our model should have no issue scaling to larger problems.} of $512 \times 512 \times 512 \times 20 \ (x \times y \times z \times t)$ when scaling the spatial dimensions of the input, and a maximum problem size of $64 \times 64 \times 64 \times 10240 \ (x \times y \times z \times t)$ when scaling the number of output timesteps. All of our scaling experiments were performed on the Perlmutter system at the National Energy Research Scientific Computing center (NERSC). Perlmutter is a flagship supercomputer, being ranked 7th in the world in terms of performance on the LINPACK benchmark as of June 2022 \citep{top500_2022}. Perlmutter consists of 1,536 GPU compute nodes, each of which contains an AMD EPYC 7763 CPU, 4x40GB A100 NVIDIA Ampere GPUs, and 256 GB of random access memory. The system utilizes a three-hop dragonfly network with Slingshot 11 interconnect fabric, allowing for data transfer up to 100 GB/s between nodes \citep{nersc2022}.

We conducted a weak scaling study (i.e. the problem size grows in tandem with the amount of computational resources) measuring the time taken to apply the domain-decomposed FNO forward without saving gradients (inference), forward with saving gradients (training), and to perform backpropagation (training). The network used is identical in structure to the original proposed by \cite{li2020fourier}, having 4 spectral convolution blocks and an embedding dimension of 20. A full table of run configurations can be found in appendix \ref{appendix:scaling}.

\begin{figure}[h]
    \centering
    \includegraphics[width=0.75\textwidth]{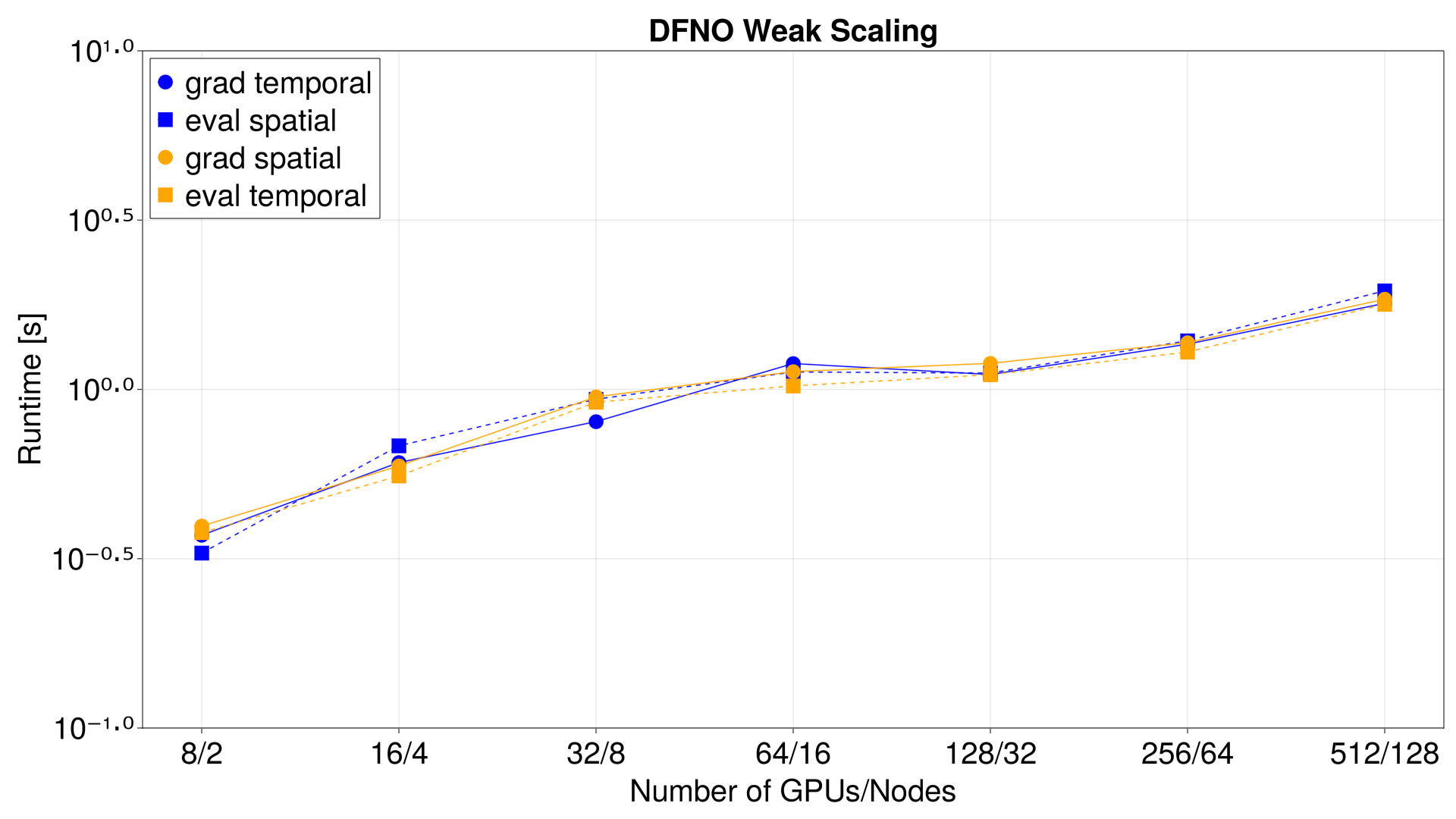}
    \caption{Weak scaling experiment results. Dashed lines indicate forward passes run with saving gradients (training scenario), and solid lines indicate forward passes run without saving gradients (inference scenario). Each run was performed using CUDA-aware MPI. We observe slightly imperfect scaling due to the nature of FNOs (i.e. all-to-all programs), but overall good performance.}
    \label{fig:scaling}
\end{figure}

\section{Conclusion}

In this work, we have presented a domain-decomposition based implementation of model-parallel Fourier neural operators for data of arbitrary size and dimensionality. Using a linear-algebraic formulation of parallelism, we derive mathematically all requisite components of distributed FNOs and provide an implementation of our distributed network in PyTorch using DistDL. We show an example of training a model-parallel 4D FNO via domain decomposition to learn solutions to the two-phase flow equations for predicting the time-evolution of subsurface CO$_2$ plumes. We demonstrate our network’s weak scaling capabilities on Perlmutter on problem sizes up to $512 \times 512 \times 512 \times 20$. To our knowledge, this is the first implementation of an FNO to scale beyond $64^3$ in the spatial dimensions. Our work provides a critical first step in the ability to solve coupled inverse and statistical problems on realistically-sized volumetric data by rapidly accelerating inference and gradient calculations on large volumetric problems via distributed operator learning.

\section*{Acknowledgments}
This research was carried out with the support of Georgia Research Alliance and partners of the ML4Seismic Center. The authors acknowledge Erik Skjetne (Equinor) and Tugrul Konuk (formerly Colorado School of Mines and Microsoft Research intern) for their contributions to the generation of the Sleipner CO$_2$ training dataset. Portions of this work were completed while Russell J. Hewett was with Virginia Tech and was funded by Department of Energy, Office of Science, Early Career Research Program award DE-SC0022041. This work is supported by the Department of Energy under Grant No. DE-SC0021515. This research used resources of the Oak Ridge Leadership Computing Facility, which is a DOE Office of Science User Facility supported under Contract DE-AC05-00OR22725. This research used resources of the National Energy Research Scientific Computing Center (NERSC), a U.S. Department of Energy Office of Science User Facility located at Lawrence Berkeley National Laboratory, operated under Contract No. DE-AC02-05CH11231 using NERSC award ASCR-ERCAP0022541.

\section*{Code Availability}
Library name: dfno

Primary Developer: Thomas Grady (SLIM Group, Georgia Tech)

Contact: \texttt{tgrady@gatech.edu}

Hardware Requirements: Multi-core or multi-GPU system

Software Requirements: MPI, CUDA (optional), Python libraries (listed at repository)

Programming Language: Python

Program Size: N/A (scripting language)

Source code: \url{https://zenodo.org/record/6463857}

\bibliographystyle{cas-model2-names}

\bibliography{refs.bib}



\clearpage
\onecolumn
\appendix
\section{Scaling Study Configurations}\label{appendix:scaling}
Table \ref{tab:runconfig} shows worker partitions and corresponding input/output sizes for spatial and temporal scaling studies. $p$ denotes the number of parallel workers, and ``Partition Shape'' denotes the Cartesian topology of those workers. Note that the given partition shape denotes only the partition of the input and output tensor of the network. Intermediate tensors within the network may have different partitions (e.g. weights in the spectral convolution or data during application of the DFFT). We note that these particular configurations were chosen in the interest of measuring our network's performance, and that a practical application would likely choose a partitioning scheme that takes into account details about the shape of its corresponding data and hardware/bandwith limitations.

\begin{table}[h]
    \footnotesize
    \centering
    \begin{tabular}{|c|c"c|c"c|c|}
        \hline
        \multicolumn{2}{|c"}{Partition Information} & \multicolumn{2}{c"}{Spatial Scaling Shapes} & \multicolumn{2}{c|}{Temporal Scaling Shapes} \\\hline
        $p$ & Partition Shape & Input Shape & Output Shape & Input Shape & Output Shape \\\hline
        1 & (1,1,1,1,1,1) & (1,1,64,64,64,1) & (1,1,64,64,64,20) & (1,1,64,64,64,1) & (1,1,64,64,64,20) \\\hline
        2 & (1,1,2,1,1,1) & (1,1,128,64,64,1) & (1,1,128,64,64,20) & (1,1,64,64,64,1) & (1,1,64,64,64,40) \\\hline
        4 & (1,1,2,2,1,1) & (1,1,128,128,64,1) & (1,1,128,128,64,20) & (1,1,64,64,64,1) & (1,1,64,64,64,80) \\\hline
        8 & (1,1,2,2,2,1) & (1,1,128,128,128,1) & (1,1,128,128,128,20) & (1,1,64,64,64,1) & (1,1,64,64,64,160) \\\hline
        16 & (1,1,4,2,2,1) & (1,1,256,128,128,1) & (1,1,256,128,128,20) & (1,1,64,64,64,1) & (1,1,64,64,64,320) \\\hline
        32 & (1,1,4,4,2,1) & (1,1,256,256,128,1) & (1,1,256,256,128,20) & (1,1,64,64,64,1) & (1,1,64,64,64,640) \\\hline
        64 & (1,1,4,4,4,1) & (1,1,256,256,256,1) & (1,1,256,256,256,20) & (1,1,64,64,64,1) & (1,1,64,64,64,1280) \\\hline
        128 & (1,1,8,4,4,1) & (1,1,512,256,256,1) & (1,1,512,256,256,20) & (1,1,64,64,64,1) & (1,1,64,64,64,2560) \\\hline
        256 & (1,1,8,8,4,1) & (1,1,512,512,256,1) & (1,1,512,512,256,20) & (1,1,64,64,64,1) & (1,1,64,64,64,5120) \\\hline
        512 & (1,1,8,8,8,1) & (1,1,512,512,512,1) & (1,1,512,512,512,20) & (1,1,64,64,64,1) & (1,1,64,64,64,10240) \\\hline
    \end{tabular}
    \caption{Scaling study run configurations for experiments performed on Perlmutter.}
    \label{tab:runconfig}
\end{table}

\end{document}